\documentclass[10pt,twocolumn,letterpaper]{article}

\usepackage{cvpr}   

\usepackage[utf8]{inputenc}
\DeclareUnicodeCharacter{2064}{}
\usepackage{listings}
\usepackage{graphicx}
\usepackage{amsmath}
\usepackage{amssymb}
\usepackage{booktabs}
\usepackage{multirow}
\usepackage{makecell}
\usepackage{soul}
\usepackage[table,dvipsnames]{xcolor}
\usepackage{changepage,threeparttable}
\usepackage[accsupp]{axessibility}
\usepackage{mmstyle}

\definecolor{MyRed}{HTML}{FF0000}

\usepackage[misc]{ifsym}

\usepackage{pifont}

\definecolor{cvprblue}{rgb}{0.21,0.49,0.74}
\usepackage[pagebackref,breaklinks,colorlinks,citecolor=cvprblue]{hyperref}

\usepackage[capitalize]{cleveref}
\crefname{section}{Sec.}{Secs.}
\Crefname{section}{Section}{Sections}
\Crefname{table}{Table}{Tables}
\crefname{table}{Tab.}{Tabs.}

\makeatletter
\DeclareRobustCommand\onedot{\futurelet\@let@token\@onedot}
\def\@onedot{\ifx\@let@token.\else.\null\fi\xspace}

\makeatother

\definecolor{turquoise}{cmyk}{0.65,0,0.1,0.3}
\definecolor{purple}{rgb}{0.65,0,0.65}
\definecolor{dark_green}{rgb}{0, 0.5, 0}
\definecolor{orange}{rgb}{0.8, 0.6, 0.2}
\definecolor{red}{rgb}{0.8, 0.2, 0.2}
\definecolor{darkred}{rgb}{0.6, 0.1, 0.05}
\definecolor{blueish}{rgb}{0.0, 0.3, .6}
\definecolor{light_gray}{rgb}{0.7, 0.7, .7}
\definecolor{pink}{rgb}{1, 0, 1}
\definecolor{greyblue}{rgb}{0.25, 0.25, 1}
\definecolor{colorFst}{HTML}{bde6cd}       %
\definecolor{colorSnd}{HTML}{e4eebc}       %
\definecolor{colorTrd}{HTML}{fff8c5}       %

\def\paperID{4} %
\def\confName{CVPR}
\def\confYear{2024}

\title{Leveraging Foundation Models via Knowledge Distillation in Multi-Object Tracking: Distilling DINOv2 Features to FairMOT}

\author{Niels G. Faber \and
Seyed Sahand Mohammadi Ziabari \and
Fatemeh Karimi Nejadasl \and
University of Amsterdam}

\begin{document}
\maketitle
\begin{abstract}
⁤Multiple Object Tracking (MOT) is a computer vision task that has been employed in a variety of sectors. ⁤⁤Some common limitations in MOT are varying object appearances, occlusions or crowded scenes. ⁤⁤To address these challenges, machine learning methods have been extensively deployed, leveraging large datasets, sophisticated models, and substantial computational resources. ⁤⁤Due to practical limitations, access to the above is not always an option. ⁤⁤However, with the recent release of foundation models by prominent AI companies, pretrained models have been trained on vast datasets and resources using state-of-the-art methods. ⁤⁤This work tries to leverage one such foundation model, called DINOv2, through using knowledge distillation. ⁤⁤The proposed method uses a teacher-student architecture, where DINOv2 is the teacher and the FairMOT backbone HRNetv2 W18 will be the student. ⁤⁤The results imply that although the proposed method shows improvements in certain scenarios, it does not consistently outperform the original FairMOT model. ⁤⁤These findings highlight the potential and limitations of applying foundation models in knowledge distillation for MOT.
\footnote{https://github.com/NissaFaber/Thesis\_repo}
\footnote{This is an MSc thesis by Niels Faber, supervised by the two other authors.}
\footnote{The original title was: Leveraging Foundation Models Using Knowledge Distillation in Multiple Object
Tracking: Using Knowledge Distillation to distil DINOv2 Features To FairMOT}
\end{abstract}
    
\section{Introduction}
\label{sec:introduction}
Multiple Object Tracking (MOT) in computer vision is a task that is associated with the detection and tracking of objects within a video sequence \cite{LUO2021103448}.  It already has numerous real-world applications such as pose estimation, visual surveillance, and behaviour analysis \cite{luo2014multiple}. In recent years, there have been many attempts to improve the state-of-the-art MOT \cite{MOT20},  focusing on overcoming  common limitations such as  'varying object appearances', 'occlusions or crowded scenes', 'data associations and identity switches'. 
To address these challenges, machine learning methods have been extensively deployed, leveraging large datasets, sophisticated models, and substantial computational resources  \cite{LUO2021103448}. 
These methods have led to improvements, however can clash somewhat with practical limitations. For example, the process of labelling data for training MOT algorithms is labour-intensive and costly. High-quality labelled datasets are essential for training robust models, but the scarcity of such data limits the potential of many machine learning approaches. In addition, advanced MOT algorithms often require substantial computational power for training and real-time inference. This requirement can be prohibitive for many applications, particularly those with limited access to high-end hardware \cite{karthik2020simple}.
Recently, several large models, or "foundation models" trained on vast datasets and extensive resources have been released by prominent companies such as Meta AI, OpenAI, and Google DeepMind \cite{bommasani2022opportunities}. The release of these models offers opportunities in many real-world applications, given that larger models offer more descriptive capabilities and can be utilized as pre-trained models for smaller, specific datasets \cite{transformerComputerVision}. However, using transformers on small datasets is not entirely straightforward. Due to the large size of foundation models, using only a small amount of data negatively impacts their performance. Some conventional approaches to address these problems are parameter-Efficient Fine-tuning (\textit{PEFT}), Low-Rank adaptation (\textit{LoRA}), or adopting \textit{distillation loss} \cite{hu2021lora,ding2023parameter}. 

Distillation loss (or knowledge distillation) involves using a foundation model as the "teacher" and a smaller model as the "student," guiding the student to replicate the performance of the teacher model\cite{knowledgeDis}. This approach has shown promising results in various machine learning applications \cite{knowledgeDis}. However, the application of knowledge distillation in multiple object tracking (MOT) using foundation models remains unexplored. This research aims to address this scientific gap by investigating the potential of using DINOv2 of \citeauthor{DINOv2} as the teacher model and FairMOT of \citeauthor{fairmot} as the student model. More specifically, an evaluation will be made on whether employing DINOv2 as a teacher model can transfer its superior feature embeddings to the FairMOT model, thereby enhancing the student's performance on public MOT datasets. By exploring this novel application of knowledge distillation in MOT, this research seeks to contribute to the field by providing insights into the effectiveness of foundation models in improving the performance of smaller, task-specific models. This led to the main research question of the thesis:

 \begin{itemize}
\item{\textit{To what extent can the foundation vision model DINOv2, utilising distillation loss, improve the performance of multiple  object tracking?}}
\end{itemize}

The resulting model will be tested on the datasets of the MOT challenge and the \textit{DanceTrack} dataset. To support  in answering this research question, the following three sub-questions were created:
\begin{itemize}
    \item How can DINOv2 feature embeddings be transferred effectively to the FairMOT model?
    \item To what degree does fine-tuning the DINOv2 model help improve the resulting model?
    \item How effectively does the proposed model generalize to a smaller private dataset?
\end{itemize}

In the following sections, the related work will be reviewed to establish the context of this study within the existing literature. Next, the methodology will detail the research design and procedures used to conduct this study. The subsequent sections will present the results, followed by a discussion of the findings, and finally, the conclusion will summarise the key insights and implications of the research. 

\section{Related Work}
\label{sec:related_work}
% Your work needs to be grounded and compared to earlier work and the state-of-the-art. Start the section with announcing the research gap and also end with the research gap. Consider using hypotheses. 

% by explaining the state of the art methods of MOT I shall show that using a foundation model to create MOT is not yet done elaborate on the concept of MOT,

In this section, the state-of-the-art MOT models will be explained, and how they differ from the aforementioned  method. Additionally, current applications and the inner workings of the DINOv2 and FairMOT models will be clarified. Finally, the common practices of Knowledge Distillation will be shown.

\subsection{Current Practices MOT}
%explanation Multiple Object Tracking
Multiple Object Tracking (MOT) is the task of locating multiple objects, maintaining their identities and tracking their trajectories, given an input video \cite{MOTreview}. Examples of input videos are pedestrians walking on the curb, and cars driving on the road. The current MOT methods are often benchmarked using the \textit{MOT17} and \textit{MOT20} datasets, as well as the DanceTrack dataset\cite{dendorfer2020mot20,sun2022dancetrack}.

Two models that score high on the MOT17 and MOT20 datasets are SMILEtrack\cite{smiletrack}, and the SparseTrack \cite{sparsetrack}. The SmileTrack approach directly addresses some key challenges in the field of MOT, such as occlusions, similar objects, and complex scenes (such as dancing people). In this algorithm, the Similarity Learning Model (SLM) is used to calculate similarities between two objects, which addresses the problem of complex scenes.  The Image Slicing Attention (ISA) blocks are an integral part of the SLM using image-slicing techniques combined with attention mechanisms for feature extraction, which are later used for the  Similarity Matching Cascade (SMC) component, that is applied for enhancing robust object matching across consecutive video frames.  The innovative approach used by the SmileTrack algorithm has helped achieve high scores on the \textit{MOT17} and \textit{MOT20} datasets.

SparseTrack is an MOT algorithm  that performs scene decomposition based on the Pseudo Depth approach  \cite{sparsetrack}. The sparse decomposition is essential for enhancing the performance of associating occluded targets.  Using pseudo-depth estimation to obtain relative depths of targets from 2D images and a depth cascading matching (DCM) algorithm that uses the obtained depth information to convert a dense target set into multiple spare target subsets and perform data association on these sparse target subsets in order from near to far. These two core elements lay the foundation of the SparseTracker which achieves comparable performances to other state-of-the-art methods on the \textit{MOT17} and \textit{MOT20} benchmarks.

The MOTRV2 model performs exceptionally well on the DanceTrack dataset. MOTRv2 improves upon MOTR, an end-to-end tracking method, by incorporating an extra object detector \cite{motrv2}. Past methods that used end-to-end object tracking methods were often outperformed by their tracking-by-detection counterparts due to their poor detection performance. Therefore, MOTRv2 is designed to enhance detection results before utilizing MOTR by formulating queries using an additional object detector to generate anchor proposals, providing a detection prior to MOTR. This modification helps with unbalanced performance between joint learning detection and association tasks in MOTR. This has in turn led to high performance on the DanceTrack dataset.

%TransTrack is a MOT method that uses Transformers 
\subsection{Foundation Models}

With the breakthrough transformer-based foundation models of Meta AI, OpenAI, and Google DeepMind, new doors opened in numerous fields of AI, with computer vision as one of them \cite{transformerComputerVision}. In computer vision, transformers have been applied to tasks including image classification, object detection, action recognition, and segmentation. The two concepts that have made transformer models such a powerful tool are self-attention mechanism, and pre-training on large corpora followed by task-specific fine-tuning \cite{liu2019roberta}. 
Self-attention permits transformers to learn relationships between elements of a sequence. Whereas traditional methods, for example, recurrent neural networks (RNN), would have to rely on recursively processing sequence elements, and can thus only attend to short-term context, the self-attention mechanism can process complete sequences and allows for learning long-term relationships \cite{attentionmodels}.
Having self-attention as the core element of transformers permits for an optimized parallelization implementation, allowing them to scale to high-complexity models and large-scale datasets. Deploying transformers on vast corpora with (un)labelled data in a (self-)supervised manner enables them to learn features in the data that can be used (without fine-tuning) to achieve performance on downstream tasks that exceeds those produced by task-specific models \cite{brown2020language}.
To optimally leverage the potential of the transformer model, it can be fine-tuned for specific tasks. This involves additional training on a smaller, task-specific dataset, adjusting the model parameters for a given task.

\subsection{DINOv2}
\begin{figure}
    \centering
    \includegraphics[scale=0.7]{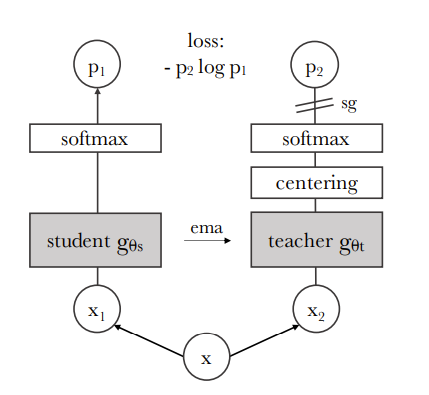}
    \caption{This figure displays the student-teacher architecture of the DINO model \cite{DINO}.  }
    \label{fig:DINOarch}
\end{figure}

DINOv2 is an example of a foundation model that uses transformers trained for image-level visual tasks (image classification, instance retrieval, video understanding) or pixel-level visual tasks (depth estimation, semantic segmentation). 
DINOv2 originated from the self-\textbf{di}stillation with \textbf{no} labels model (DINO). DINO uses a self-supervised method to learn representations of images and uses a teacher and student model to deploy knowledge distillation \cite{DINO}.
These models use vision transformers (ViT), that similar to NLP transformers, learn relations within the image data. The student and teacher models start off as an exact copy, however, during the training process, they are exposed to different variants of the same image. The data used for training is the imageNet dataset without labels. Before an image is used, a couple of augmentations are created, subdivided into global crops (more than 50\% of the original image is still in the picture) and local crops (smaller parts of the image). The teacher model is trained on the global crops, while the student model is trained on both the global and the local crops. The outcome of the teacher model is centred by computing a mean over the batch, after which both model outputs go through a heatmap-softmax. Lastly, the cross-entropy loss is calculated, to see how similar the models perform. A stop-gradient (sg) operator is applied to the teacher to propagate gradients only through the student. The teacher parameters are updated with an exponential moving average (ema) of the student parameters. The afore-explained process of self-distillation is visualised in figure \ref{fig:DINOarch}.

By using the original DINO architecture and iBOT, DINOv2 was created as an improvement of the original DINO model. IBOT uses an approach combining self-supervised learning with an online tokenizer, that adaptively learns to tokenize the input images into visual tokens directly during the training process.
Besides the novel architecture, a new data collection method was deployed, which in essence combined a larger curated dataset with uncurated data from web-crawled datasets, resulting in the LVD-142M  dataset. The modifications in the architecture and this larger dataset allowed DINOv2 to improve over DINO.

\subsection{Distillation Knowledge}
The concept of distillation knowledge has become increasingly relevant due to current state-of-the-art models becoming larger and having more parameters than ever. Models similar to this have a high computational complexity and also have large storage requirements. Knowledge distillation is essentially a model compression technique that takes large models as a teacher and smaller student models and teaches the optimal parameters settings to the student model through distillation loss \cite{knowledgeDis}. This approach is optimal when resources such as datasets, or computational power are limited.

\subsection{FairMOT}
Another MOT model that performed well on the benchmarks is the FairMOT model. At the time, most MOT models would use separate models for object detection, which is the detection of objects of interest in bounding boxes in each frame. Association, which extracts re-identification (re-ID) features from the image regions corresponding to each bounding box, links the detection to one of the existing tracks or creates a new track according to certain metrics defined on features \cite{fairmot}. FairMOT integrates both into a single framework, allowing the model to share features between detection and Re-ID, improving the efficiency and accuracy of the tracking. FairMOT is built on a backbone responsible for extracting features. These backbones are pre-trained on datasets such as the crowd-human data, COCO or imageNet. 

One of the available backbones is the HRNetV2 network. HRNetV2 is an adaptation of a deep convolutional neural network (DCNN) that maintains high-resolution representations through the whole process. It is characterised by connecting high to low convolution (or low resolution) streams in parallel and repeatedly exchange information across convolution streams. In four different stages in the network, using strided convolutions, higher resolution images are downsampled to a lower convolution stream. From the first stage, every time an extra resolution stream is created, information is exchanged between the layers. This exchange either happens through the aggregation of strided convolutions creating a new resolution stream, or aggregated upsampled convolutions to layers of a higher convolution stream. A depictive image can be found in the appendix \ref{sec:apx:hrnet}. HRNetV2 is distinguished from HRNetV1 by the final step, which upsamples all lower convolution streams to the resolution of the highest (original) convolution stream and aggregates the result \cite{hrnet}.

Upon the backbone, a detection and association branch are added. The detection branch is built on top of CenterNet and consists of three parallel heads responsible for estimating heatmaps, object centre offsets and bounding box sizes. The association branch (or Re-ID branch), aims to generate features that can distinguish objects by using  a convolution layer with 128 kernels on top of backbone features\cite{fairmot}.

Since FairMOT uses a backbone that extracts feature embedding for both the detection and re-ID, combining DINOv2, which is a model created for learning robust visual features \cite{DINOv2}, could potentially enhance the feature embeddings of the FairMOT backbone model. This, combined with the fact that FairMOT is considered a simple model \cite{fairmot}, is the reason behind using FairMOT as the student model.

\section{Methodology}
\label{sec:methodology}
% Focus on what you add to the existing method. Explain what you will do and why (and how). Do not forget to characterize your research design. There should be an evaluation plan in this section. (For DS students, this 

\begin{figure*}
    \centering
    \includegraphics[scale =0.3]{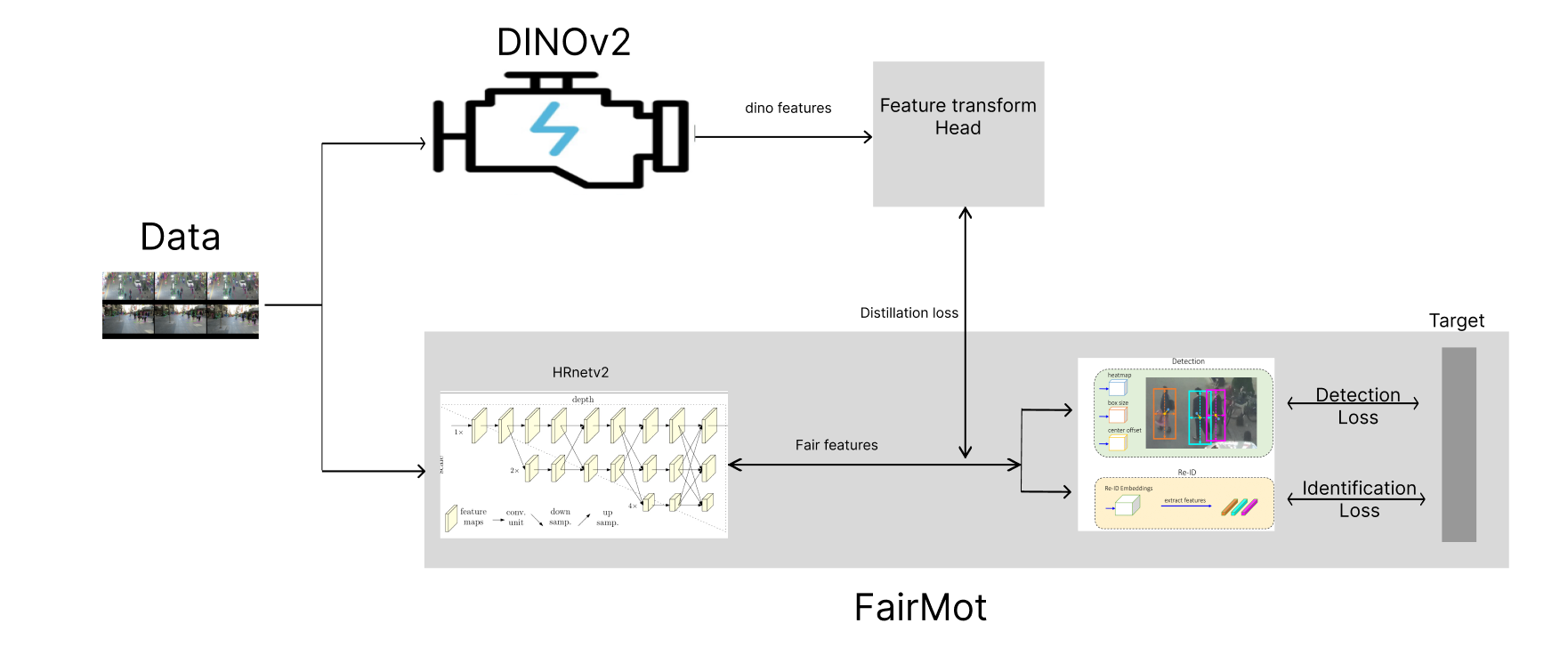}
    \caption{This figure shows the complete knowledge distillation pipeline. The data is fed in parallel to the DINOv2 model and the FairMOT model, both generating their features. The DINOv2 model returns the features from the last hidden layer. Similarly, the features of the FairMOT model are also from the last hidden layer, directly after the upsampling of the different convolution streams. Using distillation loss, the DINOv2 model teaches improved features to the FairMOT model.}
    \label{fig:pipeline}
\end{figure*}

This section outlines the approach used to address the main research question and the sub-questions. The creation and functionality of the pipeline is clarified. to answer both the main research questions and accompanying sub-questions, as well as how the research has been validated.

\subsection{Approach}
The focus of this research is on creating an effective knowledge distillation pipeline, utilizing a teacher-student model architecture. In this setup, the teacher model transfers its knowledge to the student model - specifically, the DINOv2 model acting as the teacher, with the FairMOT model being the student. DINOv2 is known for its exceptional feature extraction capabilities, enabling it to perform well in downstream tasks without fine-tuning \cite{DINOv2}.

The  HRNetV2 backbone, of the FairMOT model, also serves as a feature extractor, generating what are known as Fair Features. These features are crucial for the model’s detection and re-identification (re-ID) heads. Given that HRNetV2 is significantly smaller than DINOv2, it stands to gain considerably from the high-quality features extracted by DINOv2. The reason why the HRNetV2 model was used for the backbone was due to the available version of the DLA-34 model being too outdated, causing dependency issues with DINOv2. The HRNetV2 model offers two versions,  W32 and W18, where the first is a larger version of the model. The reason why the smaller model was used was because of computational complexity. The paper of \citeauthor{fairmot} suggested that larger backbones do not necessarily improve performance, therefore using the computational less complex model was the logical decision.  

There are various knowledge distillation methods, and this study focuses on feature-based knowledge distillation \cite{knowledgeDis} which emphasises teaching the feature embeddings (hidden states) from the teacher to the student. This approach leverages the strengths of DINOv2 in feature extraction to enhance the performance of the smaller FairMOT model.\\\\\\

\textbf{Transferring Feature Embeddings}

\subsubsection{Loss Function}

Feature distillation from teacher to student models is guided by a loss function, which measures how closely the student's features match the teacher's, and selecting the appropriate loss function is crucial. In this research, we compared two commonly used loss functions: cosine embedding loss and mean squared error (MSE) loss, and finding which of the two is more fit for feature distillation. The cosine embedding loss measures the angular difference between two flattened feature embeddings \cite{cosineloss}, while the MSE loss calculates the element-wise difference between embeddings \cite{mse}.

\subsubsection{Feature Transformation}
For the loss functions used in this research, the input tensors (feature embeddings) must have the same shape. Since the two architectures are fundamentally different, their initial output embeddings differ.
DINOv2, produces patch embeddings of shape (batch size, patches, hidden size)).

Patches are added because of the transformer's self-attention mechanism, which grows in size exponentially with the dimension of the input data. To manage this, the input image is subdivided into patches, and the resulting embeddings hold information for each patch. Learnable positional tokens are used to encode spatial information of each patch \cite{transformerComputerVision}. Additionally, a special CLS classification token is created, however is neglectable for the application of DINOv2 in this work.

HRNetV2, an adaptation of a  (DCNN), and outputs a feature map of the shape (batch size, channels, height, width)
To align the two embeddings, DINOv2’s output must be transformed from patch embeddings to spatial features ($(Batchsize,Patches,HiddenState) \rightarrow (Batchsize , Channels, Height, Witdh$)). This can be done using the Hugging Face module, where DINOv2 is set to "backbone" \cite{huggingface}. In this process, the positional embeddings are added to the patch embeddings to encode spatial information of each patch. This step is followed by the encoder part of the transformers. Simply put, the encoder processes these embeddings to capture the relationships between patches. Subsequently, the output embeddings of the encoder can be transformed to the correct dimension. To go from a 3D patch embedding to a 4D feature map, the patches are subdivided into the number of height patches and the number of width patches.  The number of height and width patches are calculated by dividing the original image height (or width) by the patch height (or width). Having calculated this, the embedding tensor can be reshaped to a 4D feature map of the shape (batch size, number of patches height, number of patches width, embedding dimension). After reshaping the embedding using a PyTorch function called '.view', which reshapes an input tensor into a given shape based on the memory location of the tensor, the reshaped tensor is permuted into the correct order: (batch size, embedding dimensions, height number of patches, width number of patches). The resulting feature map can be made compatible with the feature map of HRNetV2. A code snippet can be found in the appendix  \ref{sec:apx:patchtransform}.

Since the final model performance will depend on how well features are transferred, two transformation methods were proposed. Method one is a simple transformation, based on the paper of \citeauthor{featuretransform}, that first applies a 2d convolutional layer to increase the channels, followed by bilinear interpolating to downsample the height and the width dimension.

Method two uses a more extensive network to increase the channel size. First, a 2d convolution is applied, followed by a batch normalisation and a ReLu function. This same pattern repeats itself once and is followed by down-sampling the height and width dimensions, using bilinear interpolation. The two transformation heads can be found in the appendix \ref{subsec:apx:transformerhead}.

Batch normalisation layers are known to cause stable learning and faster convergence \cite{batchnorm}, and ReLu layers for adding non-linearity, that add stable gradients, allowing for this network to learn complex relationships in the data \cite{ReLu,distlossfunction}.
The aim of this increased complexity is to allow for a more rich representation of the features. After the two transformation methods were created, they were both tested and compared to establish which version seemed to be transferring the features most effectively. 

The direction of the transformation was initially also considered in the research, both trying the transformation of students features to the teachers and vice versa. However, after testing, the first proved more effective and given that other research in knowledge distillation suggests that the student's embeddings should be transformed to those of the teacher \cite{featuretransform}, the inverse direction was neglected in the results.

\subsubsection{Knowledge Distillation Pipeline}
Having a loss function and a method to transform the feature embeddings to a compatible shape, the final pipeline was assembled, which can be seen in figure \ref{fig:pipeline}. The data is loaded into the student and teacher model, which both extract features. Although now, where the regular FairMOT model would send the extracted features to detection and re-id head, the features are also transformed, and compared to the DINOv2 features. Hence, the detection and re-id head losses are calculated as usual, however now the distillation loss is also added to the total loss. The resulting loss function is formulated as follows:
\[
    loss = (1 - \alpha)\cdot L_{MOT} + \alpha \cdot L_{distillation}(T_{emb}, S_{emb})
\]
$L_{distillation}$ is the distillation loss function, that can be set to either cosine embedding loss or MSE loss (formulas can be found in the appendix \ref{sec:apx: distlossloss}). $L_{MOT}$ is the regular FairMOT loss function as described in the work of \citeauthor{fairmot}, and $\alpha$ is a hyperparameter between $0$ and $1$ that determines the ratio of distillation loss added to the loss function. $T_{emb}$ and $S_{emb}$ are the Teacher's and student's feature embeddings.

The $\alpha$ parameter was added, based on the paper of \citeauthor{distlossfunction}, to test whether the effect of the teacher should be emphasised or reduced in the learning process, which can in turn explain how effective the teacher's features are for the model.
\\

\textbf{Fine-tuning DINOv2}

\subsubsection{ DINOv2 sizes}
As DINOv2 is a foundation model, it already has high-performance visual features that can be directly employed with classifiers as simple as linear layers on a variety of computer vision tasks \cite{DINOv2}. The performance of the DINOv2 model is partially due to its substantial size, therefore fine-tuning such a model is a rather expensive task, and was omitted in this research. Instead, a different approach was taken. The Facebook research group released multiple sizes of the DINOv2 model, that can be distinguished on Hugging Face between small, base, large, and giant \cite{huggingface, DINOv2}. This allowed for tests to be done on the different sizes of the DINOv2 model, with the goal to find whether larger versions of DINOv2  are more beneficial for knowledge distillation. Giant was not used in this work, because this model was too large for any of the project's available GPUs.
\\

\textbf{Generalisation}

\subsubsection{Private dataset}
A crucial aspect of evaluating the effectiveness of the knowledge distillation pipeline is to assess its ability to generalise to different datasets. Particularly smaller private datasets that may not have the extensive variety and volume of data available in public datasets. The goal is to ensure that the student model, once trained using knowledge distillation, can maintain high performance and adaptability across various types of data, demonstrating its robustness and practical utility. The dataset that was used for this is the fish data set delivered by the research supervisor, and will be explained in the next section.
\\

\section{Experimental Setup}

This section will go over the data used, and the experiments run to answer both the main research questions and accompanying sub-questions, as well as how the research has been validated

\subsection{Data \& Resources}
The model was benchmarked on the following datasets: The MOT17, MOT20, and \textit{DanceTrack}. These datasets are commonly used to benchmark MOT models, making them ideal for evaluating the proposed model. As for most MOT datasets, the video sequences are split into train and test sequences. The annotation process of the tracking boxes was professionally supervised \cite{MOT16, MOT20, DanceTrack}. In the appendix \ref{sec:apx:MOTframes}, a frame of a video sequence from each dataset can be found.

\subsubsection*{MOT17 \& MOT20  }
The MOT17 dataset contains 14 public scenes, of which 7 are indoor scenes and 7 outdoor scenes, and where pedestrians are the objects of interest. MOT17 is characterised by its real-world complexity, including crowded scenes, occlusions, and various human activities. Most of the video sequences have a resolution of 1920x1080,  with a couple of exceptions that are 640x480. The FPS varies from 14 to 30, and the total number of boxes per video varies between 5325 and, 47557. the lengths of the videos lie between 20 seconds and 1 minute.

The MOT20 dataset contains 4 different sequences, that are captured in extremely crowded scenes, which makes it significantly more challenging than those in MOT17. The key challenge of MOT20 lies in its focus on very high-density crowds, making it a rigorous test for algorithms' robustness against occlusions and interactions among multiple targets. The resolutions of the videos don't vary extremely, with one video being 1654x1080, one 1173x880, and two 1920x1080. All videos have an FPS of 25. The time of these videos varies between 17s to 2:13 minutes. The amount of boxes per video lies between 26647 and 751330.

\subsubsection*{\textit{DanceTrack}} DanceTrack is a more recent MOT dataset, designed to address some of the limitations of existing datasets by providing a new challenge: tracking dancers in complex choreographies. It contains 100 videos, 40 for training, 25 for validation, and 35 for testing. DanceTrack stands out for its focus on dynamic scenarios with rapid movements and pose changes, offering a distinct set of challenges from pedestrian tracking in MOT17 and MOT20. The data consists of high-quality sequences with 20 FPS and an average length of 52.9 seconds. In total, the data consists of 105k frames and 877k high-quality bounding boxes.

\subsubsection{Fish Data}
To test how efficiently the proposed model generalizes to a private dataset, fish tracking videos provided by UvA supervisors were used. It consists of 182 videos, 84,464 frames, 3,796 tracks, 1,469,616 boxes. The sequence time varies from 2 to 14 seconds, and has a resolution of 1080 x 1920. As opposed to the benchmarking dataset, the general visibility of the objects is worse, making the task of MOT more challenging. 

\subsubsection{Resources}
Due to MOT being dependent on large amounts of video data and performing complex calculations for detection, feature extraction and tracking across multiple object frames, having access to GPUs is necessary for running the proposed MOT model. Therefore, the Snellius supercomputer and Colab Pro were used for this project. Colab Pro was mainly used for debugging, offering access to L4, T4 and A100 GPUs. All large model runs were done on the Snellius supercomputer, that came with an NVIDIA A100, 40 GB HBM2 GPU.

\subsection{Experiments}

\subsubsection{Baseline}
To understand the effect of distilling the DINOv2 features to the HRNetV2 backbone, per dataset baselines were created using the regular FairMOT model. This gives a good depiction of how well the model performs without a knowledge distillation pipeline.

\subsubsection{pre-processing}
The general layout of the MOT data is that each video sequence has one MOT annotation file. The input of the FairMOT model requires the input videos to be individual frames, where each frame has  its own annotation .txt file. The video sequences of both the MOT and DanceTrack datasets were already subdivided into single frames. The video sequences of the fish data were in mp4 format, with each their annotation file. To transform this to the correct format, all mp4 files were transformed into sequences of .jpg files. After the video sequences of all datasets existed out of .jpg files, the according annotation .txt files could be created. For this, the dataset folders should be structured as shown in the appendix \ref{sec:apx:filestruct}. The annotations were created by subdividing the MOT annotation files per video into singular frames. After this was achieved, a script was run to generate the paths to the training files and the according labels, which the 'trainloader' used to locate the training files.

\subsubsection{Ablation Study}
To construct the optimal knowledge distillation pipeline, an ablation study was conducted to find the most effective loss, transformation method, alpha setting and DINOv2 size. The effectiveness of the different methods were evaluated using the mot17 dataset. The video sequences were cut in half, using half for training and half for validation. This method is similar to that of \citeauthor{transtrack}. The reason for using a half split was to bring down the computational cost, because using only a half split reduces the training time significantly, allowing for a time-efficient ablation study.

General settings of the FairMOT model were left untouched for both the baseline FairMOT and the FairMOT section of the knowledge distillation pipeline, allowing for a more fair comparison. Hyperparameter tuning could influence the knowledge distillation pipeline and the original FairMOT differently. The setup of the ablation study can be found in Table \ref{tab:ablation}. The only settings that were changed besides the knowledge distillation settings were the batch size, set to 1, the number of workers, set to 12, and the number of epochs, set to 10. Having the batch size set to 1 was a necessity, since otherwise the DINOv2 large model could not fit into the pipeline. The models were run for 10 epochs, to avoid high variance in the results that might happen when using a low number of epochs.

\begin{table}[ht]
    \small
    \centering
    \begin{tabular}{lcccc}
    \hline
          exp. + name&  Loss&  Multi-Layer&  Alpha&  Size\\
          \hline
          1. Loss function&  Cosine&  False&  0.5&  Base\\
          &  MSE&  False&  0.5&  Base\\
          \hline
 2. Multi-layer& res. exp1& False& 0.5
&Base\\
 feature adapeter& res. exp1& True& 0.5&Base\\
 \hline
 3. Alpha& res. exp1& res. exp2& 0.25&Base\\
 & res. exp1& res. exp2& 0.50&Base\\
 & res. exp1& res. exp2& 0.75&Base\\
 \hline
 4. Size& res. exp1& res. exp2& res. exp3&Small\\
 & 
res. exp1& 
res. exp2& res. exp3&Base\\
 & res. exp1& res. exp2& res. exp3&Large\\
 \hline
    \end{tabular}
    \caption{The experiment setup of the ablation study. 'res. expX' means result of experiment X.}
    \label{tab:ablation}
\end{table}

\subsubsection{Complete Model Runs}
The models were tested on all datasets to see whether the knowledge distillation pipeline outperforms the regular FairMOT model. First, a baseline was established, by running the default FairMOT model on all datasets. Then, using the results of the ablation study, the optimal knowledge distillation model was run on all datasets. Each model was trained for 20 epochs, similar to the approach used by \citeauthor{fairmot}. The batch size was set to 4 and the number of workers to 12. Once more, the default settings were used for the FairMOT part of the pipeline.

Between the datasets, there is a slight difference in training and validation splits. Due to lack of access to the ground truth labels of the MOT17 and MOT20 dataset, the training set was split up into a training and validation set. This means that for MOT17, out of the 7 video sequences, 5 were used for training and 2 were used for validation. As for the MOT20 dataset, out of the 4 video sequences, 3 were used for training and 1 was used for validation. For the MOT17 dataset, video sequences with different annotation techniques can be chosen. For this study, the SDP method was chosen, since it was the default option in the FairMOT GitHub repository (can be found \href{https://github.com/ifzhang/FairMOT}{here}).
For the DanceTrack and fish dataset, there was access to the ground-truth labels for both training and validation. For the DanceTrack a decreased version was used to decrease the model's running time, using  a training and validation set consisting of 40 and 25 videos respectively. Finally, for the fish dataset, 110 videos were used for training and 74 for validation.

\subsection{Evaluation}
To find the effects of a foundation model on the FairMOT model, the baseline should be set to the results of the FairMOT model on the \textit{MOT17}, \textit{MOT20}, and \textit{DanceTrack} datasets. Similar metrics will be used as in the work of \citeauthor{fairmot}, which are the HOTA, CLEAR and the F1 metrics \cite{hota,clear,F1}. These are a set of metrics commonly used to evaluate the performance of MOT models.

\subsubsection*{MOTA} The Multiple-Object Tracking Accuracy (MOTA) metric combines the \textit{False Positives} (FP), \textit{False Negatives} (FN), and identity switches (ID Sw.) to encapsulate the tracking accuracy. It is calculated as follows:
\[MOTA=1- \frac{\sum_t(FN_t+FP_t+IDS_{w_t})}{\sum_t GT_t}\]
 $t$ is the index of the frame $GT$ is the number of ground truth objects.

\subsubsection*{MOTP} The Multiple-Object Tracking Precision (MOTP) measures the precision of the tracked object locations, specifically the average overlap between the predicted and the ground truth bounding boxes for correctly matched detections. It reflects the ability of the tracker to accurately localise objects, irrespective of its ability to maintain their identities over time. MOTP is given by: 

\[MOTP = \frac{\sum_{i,t}d_{i,t}}{\sum_tm_t}\]
$m_t$ denotes the number of matches in frame t and $d_{t,i}$ is the bounding box overlap of target $i$

\subsubsection*{IDF1}
IDF1 measures the consistency of object identities across frames. It is the F1 score calculated from the numbers of correctly true positives (TP), false negatives (FN), and false positive (FP). It is calculated as follows:
\[IDF1 = \frac{2\times IDTP}{2\times IDTP+IDFP+IDFN}\]
where IDTP, IDFP, and IDFN are the numbers of true positive, false positive, and false negative identifications, respectively \cite{ristani2016performance}.

\subsubsection*{MT}
The mostly tracked (MT) measures the proportion of ground truth trajectories that are tracked by the algorithm for a large majority of their presence in the video. A trajectory is considered mostly tracked if it is tracked for at least 80\% of its ground truth lifespan. First $T$ is calculated as follows:
\[\frac{\text{Length of Trajectory Tracked}}{\text{Total Length of Ground Truth Trajectory}} \geq 0.8\]
After which, MT is calculated:
\[MT = \frac{\text{Number of Mostly Tracked Trajectories}}{T} \times 100\]
\subsubsection*{ML}
 Mostly loss (ML) measures the proportion of ground truth trajectories that are seldom tracked by the algorithm. A trajectory is considered “mostly lost” if it is tracked for no more than 20\% of its ground truth lifespan. $T$ is calculated similarly as before: 
 \[\frac{\text{Length of Trajectory Tracked}}{\text{Total Length of Ground Truth Trajectory}} \leq 0.2\]
 ML is then calculated as:
 \[ML = \frac{\text{Number of Mostly Lost Trajectories}}{T} \times 100\]
The aim of these metrics is to display the strength and weaknesses of the models, allowing for a complete evaluation \cite{MOT16}.

\section{Results}
\label{sec:results}
This section will show the results of the different research questions. Starting off with the ablation study experiments, to see how the features are best distilled and how the teacher's model size impacts the performance. Followed by the results on the complete datasets.

\subsection{Ablation Study}

The experiments conducted in this research aimed to determine the most effective loss function and feature transformation method for the knowledge distillation pipeline, using the MOT17 dataset as the evaluation benchmark.
\\

\subsubsection{Loss Function Evaluation}

The result of the loss function evaluation experiment can be found in  Table  \ref{tab:loss} below. The results of this experiment indicate that the cosine embedding loss outperformed the MSE loss across several metrics.

\begin{table}%[H]
    \centering
    \small
    \begin{tabular}{|c|c|c|c|c|c|} \hline 
         Loss Function/Metric & MOTA & MOTP & IDF1 & MT & ML \\ \hline 
         Cosine & 50.0\% & 0.257 & 60.0\% & 63 & 107 \\ \hline 
         MSE & 48.0\% & 0.260 & 57.7\% & 61 & 113 \\ \hline
    \end{tabular}
    \caption{This table shows the results of the cosine embedding loss and the MSE loss. }
    \label{tab:loss}
\end{table}

From Table \ref{tab:loss}, it is evident that the cosine embedding loss achieved a 2\% higher MOTA and a 2.3\% higher IDF1 score compared to the MSE loss. The MOTP, MT, and ML metrics were relatively even between the two loss functions. Based on these results, the cosine embedding loss was selected for subsequent experiments due to its superior performance.
\\
\subsubsection{Feature Transformation Evaluation}

The results of the 'feature transformation method' experiments are presented in Table \ref{tab:multil_layer}.

\begin{table}%[H]
    \centering
    \begin{tabular}{|c|c|c|c|c|c|} \hline 
         Method/Metric& MOTA & MOTP & IDF1 & MT & ML \\ \hline 
         Single-Layer& 50.0\%& 0.257 & 60.0\% & 63 & 107 \\ \hline
         Multi-Layered & 47.8\% & 0.246 & 58.6\% & 48 & 131 \\ \hline
    \end{tabular}
    \caption{This table shows the results of the different transformation methods used.}
    \label{tab:multil_layer}
\end{table}

The results in Table \ref{tab:multil_layer} show that the simpler single-layer transformation outperformed the more complex multi-layered transformation. The single-layer method achieved higher scores in MOTA and IDF1, as well as a better balance in MT and ML metrics, suggesting that simplicity in the transformation process can lead to more effective feature transfer.

\subsubsection{Alpha Parameter Tuning}

The results of varying alpha values are shown in Table \ref{tab:alpha}. From the Table, it is clear that an alpha value of 0.50 resulted in the best overall performance, achieving the highest MOTA and IDF1 scores. Lower (0.25) and higher (0.75) values of alpha led to a decrease in performance, indicating that a balanced weighing of the distillation loss is crucial for optimal knowledge transfer.

\begin{table}%[H]
    \centering
    \begin{tabular}{|c|c|c|c|c|c|} \hline 
         Alpha& MOTA & MOTP & IDF1 & MT & ML \\ \hline 
         0.25 & 49.9\% & 0.246 & 60.5\% & 61 & 118 \\ \hline 
         0.50 & 50.0\% & 0.257 & 60.0\% & 63 & 107 \\ \hline 
         0.75 & 43.4\% & 0.244 & 54.9\% & 54 & 133 \\ \hline
    \end{tabular}
    \caption{This table shows the results of the different $\alpha$ parameter values.}
    \label{tab:alpha}
\end{table}

\subsubsection{DINOv2 size}

The effectiveness of the knowledge distillation process can also be influenced by the size of the teacher model. To investigate this, different sizes of the DINOv2 model were tested to determine their impact on the student model's performance. The results for the different model sizes—Small, Base, and Large—are summarized in Table \ref{tab:size}.
\begin{table}%[H]
    \centering
    \begin{tabular}{|c|c|c|c|c|c|} \hline 
         Size&  MOTA&  MOTP&  IDF1&  MT& ML\\ \hline 
         Small&  48.4\%&  0.249&  59.6\%&  51& 132\\ \hline 
         Base&  50.0\%&  0.257&  60.0\%&  63& 107\\ \hline 
         Large&  51.1\%&  0.256&  59.3\%&  60& 107\\ \hline
    \end{tabular}
    \caption{This table shows the results of the different DINOv2 model sizes.}
    \label{tab:size}
\end{table}

Based on these results, it can be inferred that the larger the teacher model, the better the overall performance in terms of MOTA. However, the Base model provided a good balance across all metrics, making it a suitable choice for knowledge distillation without the additional computational costs associated with the Large model.

\subsubsection{Results Ablation Study}
From Tables \ref{tab:loss}, \ref{tab:multil_layer}, \ref{tab:alpha} and \ref{tab:size} the optimal model settings for the complete datasets were taken. Hence, for the model's validation on the complete data, the following methods were used:
\begin{itemize}
    \item Cosine embedding loss
    \item Single-layer transformation
    \item $\alpha = 0.50$
    \item DINOv2 size: Base
\end{itemize}

These experiments highlight the importance of selecting appropriate loss functions, feature transformation methods, and hyperparameter values in the knowledge distillation process. The cosine embedding loss, simpler feature transformation, a balanced alpha value of 0.50, and the Base DINOv2 model were found to be the most effective in enhancing the performance of the student model on the MOT17 dataset.

\subsection{Complete Dataset}
Using the results from the ablation study, the complete knowledge distillation pipeline was constructed and validated on all datasets. The results of the knowledge distillation pipeline were compared to the baseline set by the regular FairMOT model and can be found in the table below.

\begin{table}%[H]
    \centering
    \small
    \begin{tabular}{lcccccc} 
    \hline
          Dataset&Method&  MOTA&  MOTP&  IDF1&  MT& ML\\ 
          \hline
          MOT17&baseline&  81.8\%&  0.200&  82.9\%&  120& 12\\
          
 & KD& 84.7\%& 0.186& 86.5\%& 132&8\\
 \hline
          MOT20&baseline
&  71.5\%&  0.242&  64.0\%&  854& 77\\
 & KD& 69.6\%& 0.252& 61.9\%& 892&68\\ 
 \hline
          DanceTrack&baseline
&  65.4\%&  0.231&  34.9\%&  89& 29\\
 & KD&       65.6\%& 0.247& 33.1\%& 118&13\\ 
 \hline

          Fish&baseline
&  27.3\%&  0.287&  41.8\%&  173& 487\\ 
 & KD& 25.6\%& 0.285& 39.4\%& 167&537\\
 \hline
    \end{tabular}
    \caption{This table shows the results of the baseline model and the knowledge distillation pipeline (KD) on the MOT17, MOT20, DanceTrack and fish datasets. In the appendix  \ref{sec:apx:RunningTimes} the number of the extension of this table can be found with the running time per epoch and the epoch returning the highest score.}
    \label{tab:final results}
\end{table}

The results indicate that applying knowledge distillation to the FairMOT model does increase performance in certain aspects. The largest improvements are made on the MOT17 dataset. Here the MOTA results improve by $2.9\%$, the IDF1 by $3.6\%$  and additionally a slight improvement in MT (by 12) and ML (4) is perceived.
The effectiveness of knowledge distillation seems less on the MOT20 dataset. Improvements in the MOTP by 0.010, MT by 38, and ML by 9 can be found. However, in terms of MOTA and IDF1 the baseline outperforms the knowledge distillation methods, by $1.9\%$ and by $2.1\%$. 

For the DanceTrack dataset, the knowledge distillation method seems to perform marginally better across 4 of the 5 metrics. The MOTA is improved by $0.1\%$, MOTP by 0.016, MT by 29 and ML by 16. The baseline still outperforms the knowledge distillation method on the IDF1 metric by $1.8\%$. 

Finally, for the fish dataset, the baseline model seems to outperform the knowledge distillation pipeline on all metrics except MOTP, which is improved by $0.003$. For the MOTA and IDF1 to perform worse on the knowledge distillation pipeline is mostly consistent with the other results. However, a drastic decrease in performance on the MT and ML is remarkable, given that on previous datasets it consistently outperformed the original model.
\section{Discussion}
\label{sec:discussion}
% Compare your results with the state-of-the-art and reflect upon the results and limitations of the study. You can already hint at future work to which you come back in the conclusion section.

In this section, the results are reflected upon, a comparison to the state-of-the-art is made, and the limitations of the study are highlighted.

\subsection{Ablation study results}

\subsubsection{Effectiveness of DINOv2 Feature Embeddings Transfer} 
In the task of effectively transferring features from teacher to student, the cosine embeddings loss function outperforms the MSE loss. The reason might lie in the fundamental difference between the two functions. As mentioned in the methodology, where the cosine embedding loss is set to calculate the angular difference, the MSE loss calculates the difference of each element. How the cosine loss might benefit from calculating angular differences is that it is good at preserving the relative orientation of feature vectors. Therefore, keeping the embedded semantic values possibly intact. In addition to this, the cosine embedding loss is generally insensitive to magnitude, this might indicate the magnitude of the features is not as important as the angular difference \cite{cossim}.

For the feature transformation method, the one-layered feature transformation proved more effective as a feature alignment tool. This indicates that a straightforward transformation process may be sufficient for effective feature transfer, avoiding the potential overfitting or increased computational complexity associated with more intricate transformations. The one-layered method similar to \citeauthor{featuretransform} method has thus proven to be effective.  

As for the $\alpha$ parameter, setting it to 0.50 resulted in the best performance, showing that in this case, balancing the MOT loss and the distillation loss is crucial for improved performance. The work of \cite{balanceknowledge} suggests that setting a weight parameter can help balance the distillation loss, ensuring the other tasks of the model are not neglected. It can be observed from Table \ref{tab:alpha} that the difference between $\alpha  =0.25 $ and  $\alpha =0.50$ is remarkably smaller than the difference between  $\alpha= 0.50$ and $\alpha=0.75$. This suggests that, if the MOT tasks of the model are neglected, it will suffer a decrease in performance.  

 \subsubsection{Impact of Model Size}
In terms of the DINOv2 model size, the results indicate that the Large model generally performs well, achieving the highest MOTA (Table \ref{tab:size}). However, since the difference with the base model is so small, and the base model is computationally more efficient whilst still outperforming the Small model on every metric, the Base model was the most suitable choice. The insignificant difference between the Base and Large model can be explained by the research of \citeauthor{fairmot}. This concluded, by trying different backbone sizes for the FairMOT model, larger backbones did not significantly increase performance. This might therefore also be valid for larger teacher models.

\subsection{Result datasets}
The results on the MOT17 dataset indicate that the knowledge distillation pipeline outperforms the original model on every aspect except MOTP, meaning that the model's precision of tracked object locations is lower. The video sequences of MOT17 are of pedestrians walking, and are usually not too crowded, which makes it a relatively less complex dataset. The MOTA on both the knowledge distillation pipeline and the baseline model are higher than the model's benchmarks, which is undoubtedly caused due to not having access to the test set \cite{fairmot}. Another remark on why the knowledge distillation pipeline scores this well, is due to the model settings and distillation methods being optimised on this dataset.

For the MOT20 dataset, the knowledge distillation does not generally outperform the baseline model. The MOTA is significantly lower, indicating that the tracking accuracy is worse in the knowledge distillation pipeline. The IDF1 is also substantially lower, suggesting that the consistency of the IDs is quite low across the frames. Both the baseline and the knowledge distillation scores were worse than on the MOT17 dataset. This can be explained by the high crowd density of the dataset, making maintaining the ID, and correctly tracking them considerably difficult. This is also shown in the results of the work of \citeauthor{fairmot}

The performance on the DanceTrack dataset of the knowledge distillation pipeline is close to that of the baseline model. The knowledge distillation pipeline scores better on nearly every metric. However, the performance on the MOTA only differs by $0.1\%$ and the IDF1 score by $1.8\%$ which is worse than on the original model. The DanceTrack dataset addresses the shortcomings of the MOT17 and MOT20 dataset \cite{sun2022dancetrack}, which are dynamic scenarios with rapid movement and pose changes. Compared to the MOT20 and MOT17 datasets, the performance is considerably worse on both baseline and knowledge distillation. This means that the FairMOT model generally is not robust against the added complexities of the DanceTrack dataset.

The last results were of the private fish dataset. This dataset was to test how well the model would generalise to a smaller private dataset. The fish dataset generally had low visibility and targets were much more similar and smaller compared to the datasets before. Therefore, a substantial drop in performance in regard to MOTA and IDF1 can be seen. The original FairMOT outperforms the distillation knowledge in all aspects. As mentioned in the results section, it seems inconsistent with the MT and ML scores of the previous datasets, in which the knowledge distillation pipeline always outperformed the original model.

A remark that should be made on the proposed method: while the knowledge distillation pipeline does not seem to outperform the original models in regard to MOTA and IDF1, it does seem to generally have superior performance on the MT and ML metrics (except for the fish dataset). This does indicate that it has slightly improved over the tracking abilities of the original models.

Given that the knowledge distillation pipeline does not generally outperform the FairMOT model, it is safe to say that the proposed method does not compete with the current state-of-the-art. The FairMOT model itself is already relatively outdated, given that numerous models score better performance on the benchmarking dataset \cite{MOT16}.

\subsection{Limitations}

\subsubsection{Model settings}
An ablation study was conducted to find the best methods and settings for the knowledge distillation pipeline. A potential limitation in the result is the fact that this study has only been conducted on the mot17 dataset. The reason for this was that the mot17 dataset is relatively small, meaning the ablation study would not be very expensive given the limited resources. It is therefore possible that if each set had undergone the ablation study, different and perhaps more optimal settings could have been found for the different datasets. In terms of validity, it is possible that the optimal results have not been found through this method. 

\subsubsection{MOT17 \& MOT20}
As mentioned in the methodology, there was no access to the ground truth labels of the test sets of both the MOT17 and MOT20 datasets for this research. Therefore, the train validation split had to be done on what was originally meant as just the training set. This is one of the reasons why the results on these datasets differ from those of the SOTA and the original FairMOT paper. Access to ground truths of the test set is needed for a more accurate representation of how the model compares to the SOTA.

\subsubsection{Backbone FairMOT}
In this work, the HRnet18 backbone was used instead of the DLA-34 backbone. This was done due to the code for DLA-34 being outdated, and therefore incompatible with the DINOv2 model. Since the DLA-34 is the backbone model that performs the best for FairMOT in the paper of \citeauthor{fairmot}, it is possible that through using the hrnetv18 backbone the optimal results were not found.

\subsubsection{General DINOv2}
A fundamental difference in the pipeline of HRNetV2 W18 backbone and the DINOv2 model is the updating of parameters. Both HRNetV2 W18 and DINOv2 are initially feature extraction models, and according to their initial state, will return certain features given an input. However, once the training commences, the weights in the HRNetV2 W18 backbone are updated, allowing for more compatible features with the MOT task. The weights of the DINOv2 models remain unchanged through the whole training process, possibly making the features of DINOv2 too general for the MOT task. 
\section{Conclusion}
\label{sec:conclusion}
% Answer each research question and address how the limitations of the study qualify the conclusion.

This work has focused on addressing the scientific gap of using foundation models through knowledge distillation in multiple object tracking. To answer this, three sub-questions were devised: How can DINOv2 feature embeddings be transferred effectively to the FairMot model? To what degree does fine-tuning the DINOv2 model help improve the resulting model? How effectively does the proposed model generalize to a smaller private dataset? 

Based on the method that was used in this research, it was found that features embeddings were transferred most effectively using the cosine embedding loss function, a single-layer transformation and setting $\alpha$ to a balanced $0.50$. 

As previously mentioned, since task specific fine-tuning was beyond the resources and scope of this thesis, different DINOv2 sizes were trialled. It was found that the dinvo2 base model returned the best performing results, based on the metrics used and its relatively limited computational complexity.
From the results on the private fish dataset, it can be concluded that the model generalises no better than the original FairMOT model itself, for the original FairMOT model scores higher than the knowledge distillation pipeline. 

Based on the results of the sub-questions and the results of the model's performances on the complete dataset (table \ref{tab:final results}) the research question:\textit{ To what extent can the foundation vision models DINOv2, utilizing distillation loss, improve the performance of multiple object tracking ?}  can now be answered. It can be concluded that the extent of improvement is quite diverse per dataset. For MOT17, the performance seems to exceed that of the original model, however as mentioned in the limitation, the MOT17 dataset was not complete due to the test set being unavailable. Besides this, it was used for the ablation study, possibly making the settings more favourable for this dataset. On the other datasets, in terms of MOTA and IDF1, the original model seemed to outperform the proposed method, while on the DanceTrack and MOT20 datasets, the MOTP, MT and ML were better compared to the original model. As for the comparison to the SOTA, since the FairMOT model is not outperformed for the most part, the influence on the SOTA will be minimal, given that there are numerous models that outperform FairMOT by a large margin \cite{MOT16,MOT20,sun2022dancetrack}. 

\section{Future Work}
A few follow-up methods could be tested to make utilising foundation models through knowledge distillation more relevant. If possible, it could be applied to a better performing MOT model than FairMOT giving it a higher starting performance, which would making an improvement over the SOTA more feasible.  Another method that could potentially improve the application of DINOv2, is task specific fine-tuning. As previously mentioned as a possible limitation, is that DINOv2 features don't adjust to the MOT task. Therefore, it might be necessary to fine-tune the DINOv2 in order for it to produce more applicable features for MOT.

{
    \small
    \bibliographystyle{ieeenat_fullname}
    \bibliography{main}
}
% You can choose whether you prefer a single or double column appendix.
% Whatever you choose, you will need to stick to it throughout the appendix.
% For double column style, comment the next line.
\onecolumn

\appendix
%\begin{appendices}

\section{Appendix}
\label{sec:apx:first_appendix}

\subsection{HRNetv2 Architecture}
\label{sec:apx:hrnet}
\begin{figure}%[H]
    \centering
    \includegraphics[scale=0.5]{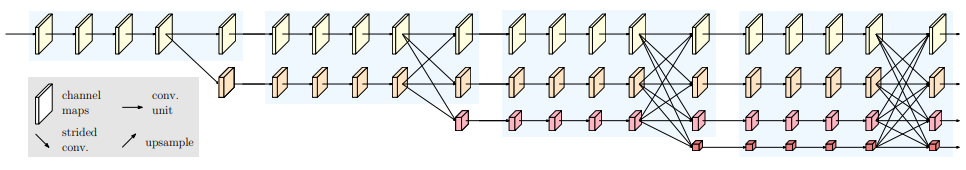}
    \caption{The architecture of the HRNet. As can be seen, the high resolution input stream is subdivided four times, into varying, from high to low, convolution streams. And at these divisions, information exchange through either aggregated strided convolutions or aggregated upsampled convolutions can be seen \cite{hrnet}.  }
    \label{fig:hrnet}
\end{figure}

\subsection{Transformation of Patch Embeddings}
\label{sec:apx:patchtransform}
This is a simplified code snippet from the Hugging Face dinvo2 transformer module (can be found \href{https://github.com/huggingface/transformers/blob/main/src/transformers/models/DINOv2/modeling_DINOv2.py}{here}). The code below belongs to the DINOv2 backbone class.
\begin{lstlisting}
# Remove the [CLS] token 
patch_embeddings = hidden_states[:, 1:, :] # Shape:(batch_size, num_patches, embedding_dim) 

# Calculate number of patches in height and width 
num_patches_height = image_height // patch_height 
num_patches_width = image_width // patch_width 

# Reshape the sequence to match spatial dimensions batch_size, num_patches, embedding_dim = patch_embeddings.shape patch_embeddings = patch_embeddings.view(batch_size, num_patches_height, num_patches_width, embedding_dim) 

# Permute dimensions to match (batch_size, embedding_dim, height, width) 

spatial_features = patch_embeddings.permute(0, 3, 1, 2).contiguous() 
\end{lstlisting}

\subsection{Transformer Head}
\label{subsec:apx:transformerhead}
\begin{figure}%[H]
    \centering
    \includegraphics[scale=0.5]{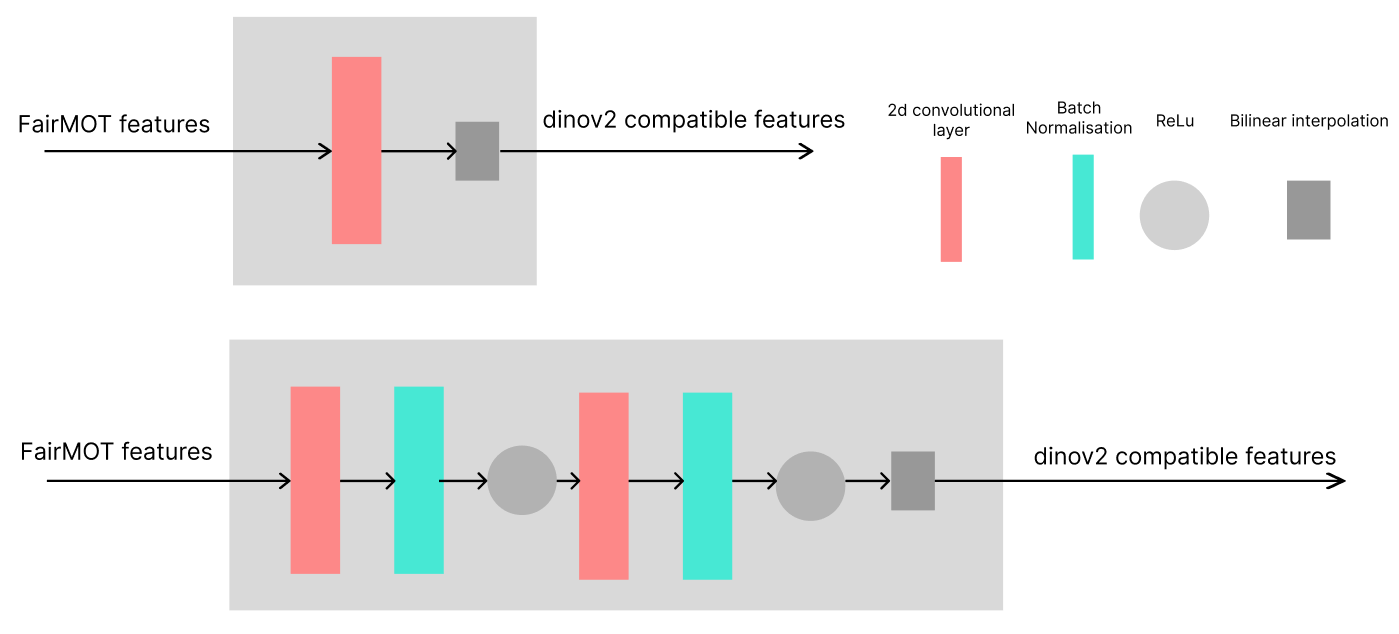}
    \caption{This image displays the two proposed transformation methods. The top one is the simple single-layered transformation, while the one on the bottom is the complex multi-layered transformation.}
    \label{fig:transformerHeads}
\end{figure}

\subsection{Distillation Loss Function}
\label{sec:apx: distlossloss}
Both loss functions are from a built-in PyTorch function \cite{pytroch}.
\subsubsection{MSE loss}
\[MSE = \frac{1}{N}\sum^N_i(Y_i - X_i)^2 \]

N is the number of samples

\subsubsection{Cosine Embeddings Loss (For maximizing similarity)}
\[loss(x,y) = 1 - cos(x_1, x_2)\]

\subsection{File Structure}
\label{sec:apx:filestruct}

The figure below displays how the files should be structured when making the data ready for training.
\begin{figure}%[H]
    \centering
    \includegraphics[scale=0.6]{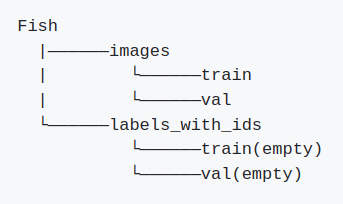}
    \caption{Example of how to structure the dataset files.}
    \label{fig:enter-label}
\end{figure}

\subsection{Train and Validation Sizes}

Table \ref{tab:trainSizes} displays the sizes of the training and validation sets, including the number of videos, frames and size in GB.

\begin{table}%[H]
    \centering
    \begin{tabular}{|c|l|c|c|l|c|c|}
    \hline
         &   Videos in train set&Train size&  Total frames train&   Videos in validation set&Validation size& Total frames validation\\
          \hline
         mot17&   5&0.578GB&  3666&   2&0.298GB& 1650\\
          \hline
         mot20&   3&1.8GB&  5605&   1&1.4GB& 3315\\
          \hline
         dancetrack&   40&6.7GB&  40887&   25&4.1GB& 25558\\
          \hline
         fish&   110&27GB&  54205&   74&14GB& 30668\\
          \hline
    \end{tabular}
    \caption{This table shows the sizes of the training and validation sets in GB's and the number of frames per set.}
    \label{tab:trainSizes}
\end{table}

\subsection{Running Times Model}

\label{sec:apx:RunningTimes}
Below is a table displaying the running times of different model runs on a Snellius A100 40GB GPU.

\begin{table}%[H]
    \centering
    \begin{tabular}{lccl} 
    \hline
          Dataset&Method&  Time per epoch (minutes)&Optimal epoch\\ 
          \hline
          MOT17&baseline&  7.416667 &20\\
          
 & KD& 20.033333 &20\\
 \hline
          MOT20&baseline
&  4.600000 &20\\
 & KD& 21.416667 &20\\ 
 \hline
          DanceTrack&baseline
&  17.233333 &10\\
 & KD&       79.600000 &10\\ 
 \hline

          Fish&baseline
&  44.466667 &15\\ 
 & KD& 176.433333 &15\\
 \hline
    \end{tabular}
    \caption{This table shows the running times and the optimal number of epochs of the baseline model and the knowledge distillation pipeline (KD) on the MOT17, MOT20, DanceTrack and fish dataset.}
    \label{tab:final results}
\end{table}

\subsection{Frames of Different MOT Datasets}
\label{sec:apx:MOTframes}

\begin{figure}%[H]
    \centering
    \includegraphics[scale=0.1]{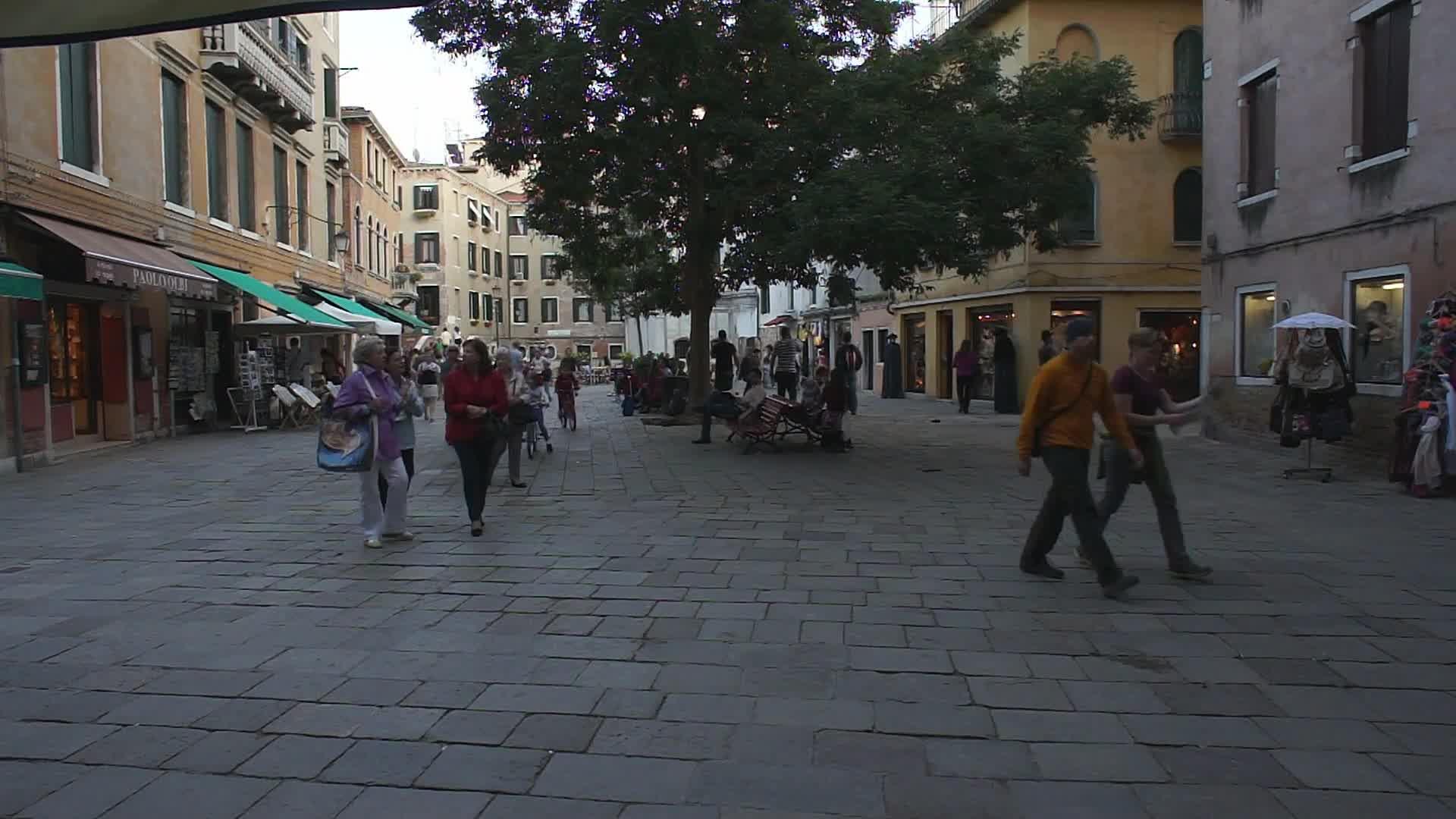}
    \caption{A frame of the one of the mot17 video sequences. A couple of pedestrians can be observed, and an all round not a too crowded scene}
    \label{fig:enter-label}
\end{figure}

\begin{figure}%[H]
    \centering
    \includegraphics[scale=0.1]{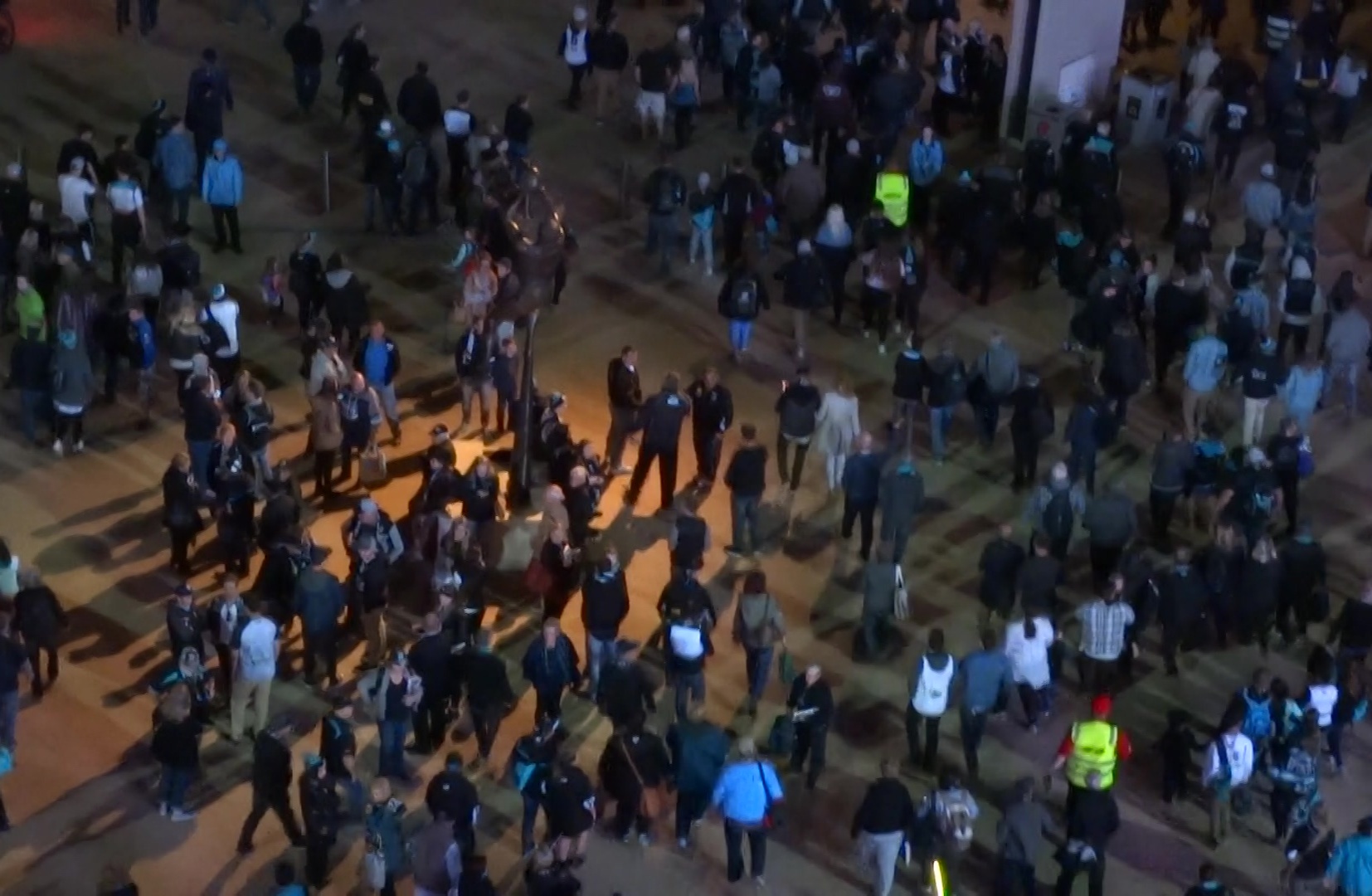}
    \caption{A frame of one of the mot20 sequences. This frame shows a very crowded scene, making an MOT task more complex.}
    \label{fig:enter-label}
\end{figure}

\begin{figure}%[H]
    \centering
    \includegraphics[scale=0.2]{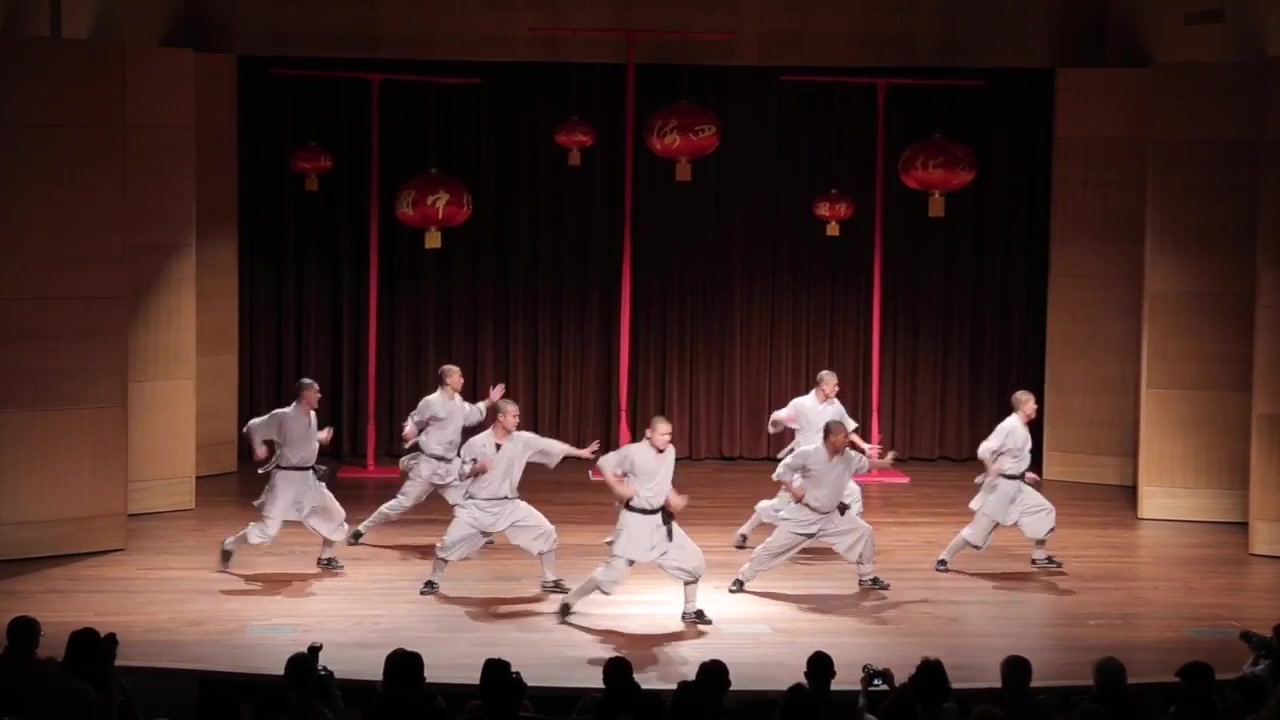}
    \caption{A frame of one of the DanceTrack sequences. This frame shows one of the DanceTrack choreographies, which consists of complex and irregular motions.}
    \label{fig:enter-label}
\end{figure}

\begin{figure}%[H]
    \centering
    \includegraphics[scale=0.2]{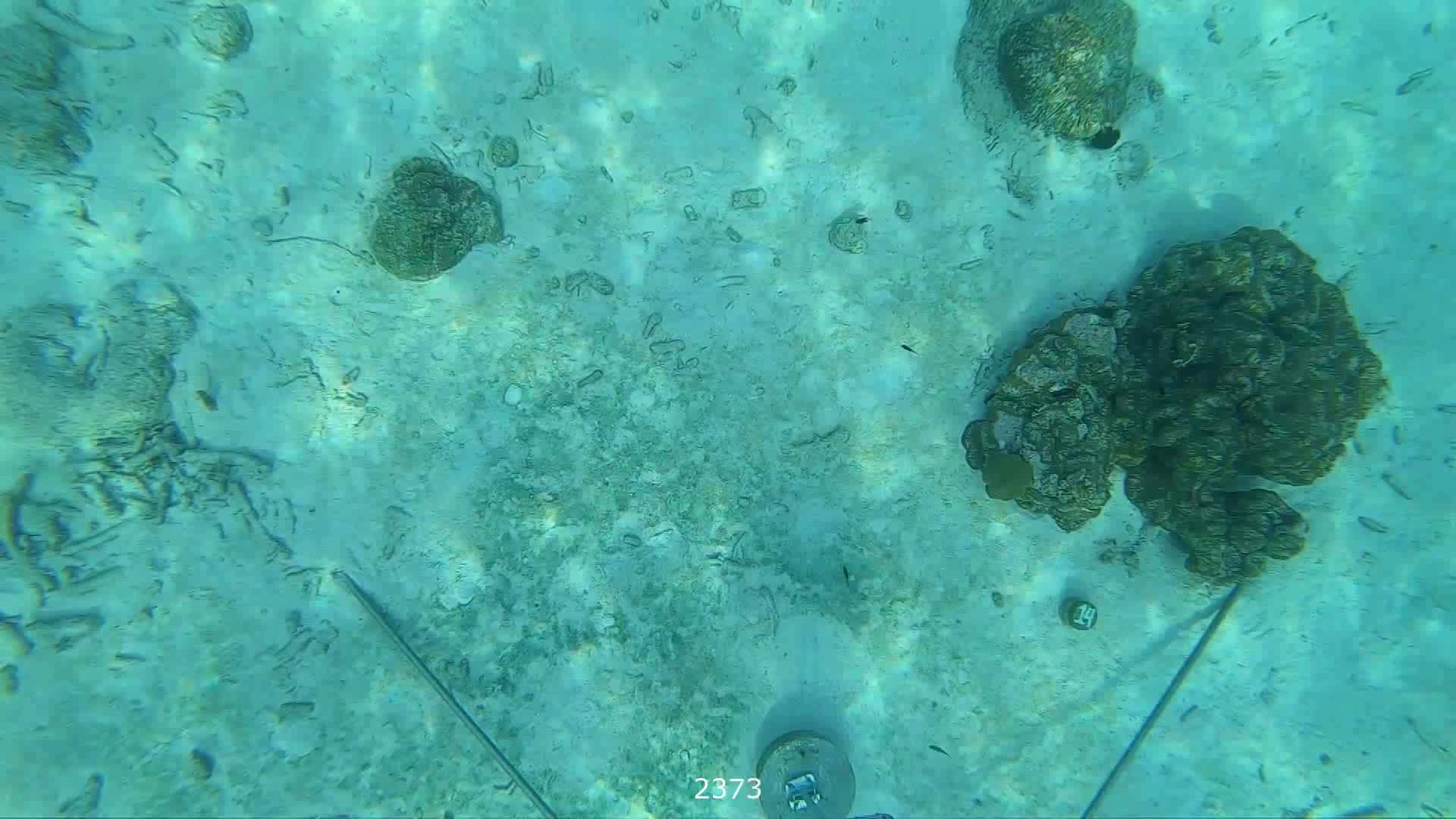}
    \caption{A frame of one of the fish video sequences. The fishes are quite small and therefore hard to perceive, making the MOT task quite difficult.}
    \label{fig:enter-label}
\end{figure}

%\end{appendices}

\end{document}